\documentclass[conference]{IEEEtran}
\IEEEoverridecommandlockouts
\usepackage{natbib}  
\usepackage{amsmath,amssymb,amsfonts}
\usepackage{algorithmic}
\usepackage{graphicx}
\usepackage{textcomp}
\usepackage{caption}
\usepackage{subcaption}
\usepackage{xcolor}
\usepackage{tabularx}
\usepackage{array}    
\usepackage{booktabs} 
\usepackage{multirow}
\usepackage{float}    

\usepackage{makecell,siunitx}
\usepackage[table]{xcolor} 
\renewcommand{\arraystretch}{1.2}
\newcolumntype{Y}{>{\centering\arraybackslash}X}
\newcolumntype{L}{>{\raggedright\arraybackslash}X}
\DeclareSIUnit\px{px}
\newcommand{\FigImageWorkflow}{%
\begin{figure}[htbp]
    \centerline{\includegraphics[scale=0.65]{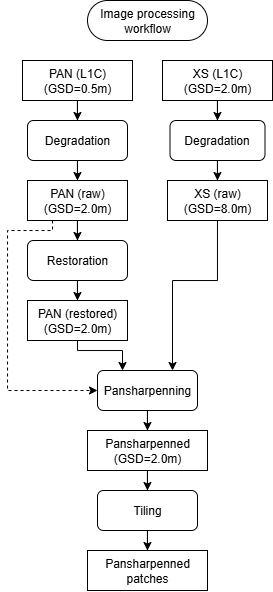}}
    \caption{Image processing workflow}
    \label{fig:image_workflow}
\end{figure} 
}

\newcommand{\FigSensorSimuFlowchart}{%
\begin{figure}[htbp]
    \centerline{\includegraphics[scale=0.58]{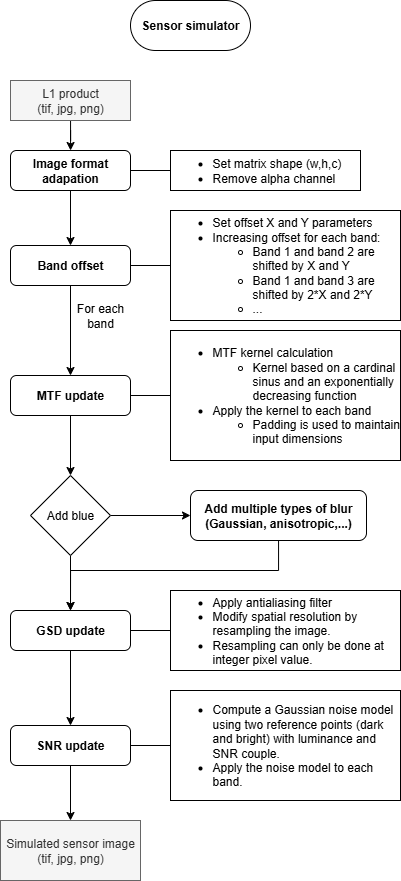}}
    \caption{Sensor simulator flowchart}
    \label{fig:sensor_simu_flowchart}
\end{figure} 
}

\newcommand{\FigSimu}{%
\begin{figure}
    \begin{table}[H]
        \centering
        \renewcommand{\arraystretch}{1} 
        \setlength{\tabcolsep}{1.5pt} 
        \begin{tabular}{|c|c|c|}
            \hline
            \centering
            \begin{tabular}{c} 
            \\
                \includegraphics[width=2.5cm,height=2.5cm]{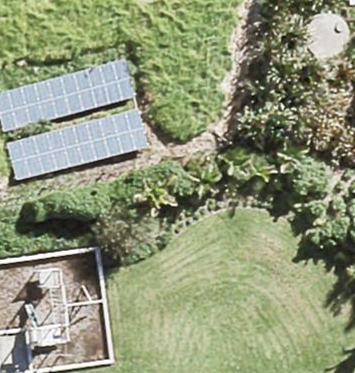} \\ 
                Reference 
            \end{tabular} & 
            \begin{tabular}{c} 
             \\
                \includegraphics[width=2.5cm,height=2.5cm]{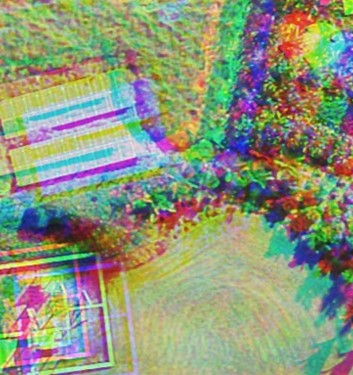} \\ 
                Band offset
            \end{tabular} & 
            \begin{tabular}{c} 
             \\
                \includegraphics[width=2.5cm,height=2.5cm]{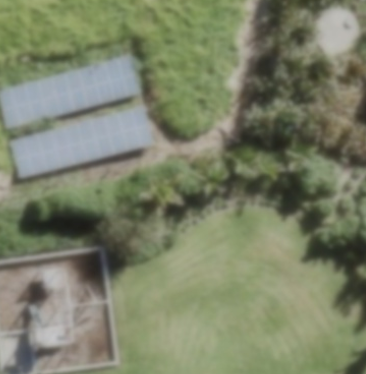} \\ 
                MTF update
            \end{tabular} \\
            \hline
            \begin{tabular}{c} 
             \\
                \includegraphics[width=2.5cm,height=2.5cm]{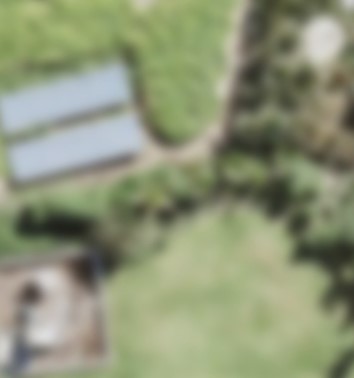} \\ 
                Add blur
            \end{tabular} & 
            \begin{tabular}{c} 
             \\
                \includegraphics[width=2.5cm,height=2.5cm]{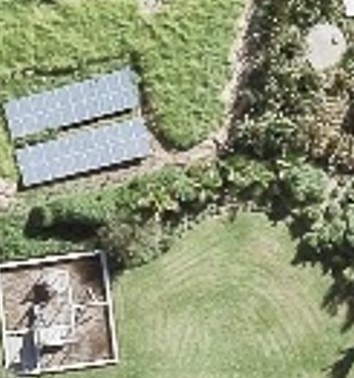} \\ 
                GSD update 
            \end{tabular} & 
            \begin{tabular}{c} 
             \\
                \includegraphics[width=2.5cm,height=2.5cm]{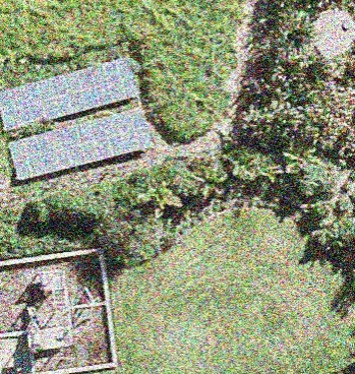} \\ 
                SNR update
            \end{tabular} \\
            \hline
        \end{tabular}
    \end{table}
    \caption{Sensor simulation demonstration on a Maxar land product. \textit{Maxar Products © 2011-2024 Maxar Technologies.}}
    \label{fig:simu}
    \end{figure}
}

\newcommand{\FigDataCompare}{%
\begin{figure}[!tbp]
     \begin{subfigure}[b]{0.32\columnwidth}
     \centering
        \includegraphics[width=\textwidth]{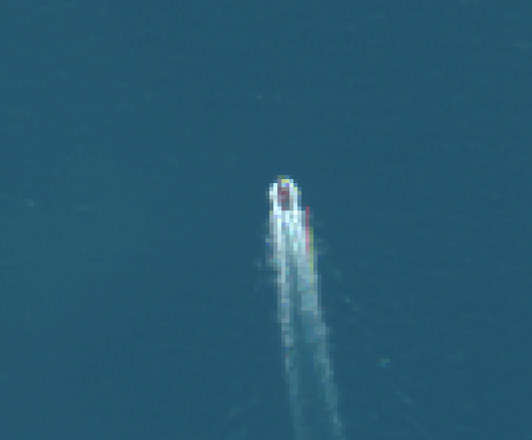}
        \caption{Reference L1 product}
        \label{fig:xs_granule}
      \end{subfigure}
    \hfill
    \begin{subfigure}[b]{0.32\columnwidth}
     \centering            \includegraphics[width=\textwidth]{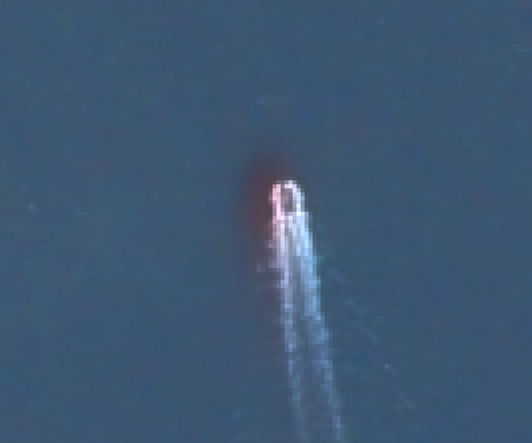}
        \caption{Simulated $\tilde{L1}$ product}
        \label{fig:l1_granule}
      \end{subfigure}
      \hfill
      \begin{subfigure}[b]{0.32\columnwidth}
        \includegraphics[width=\textwidth]{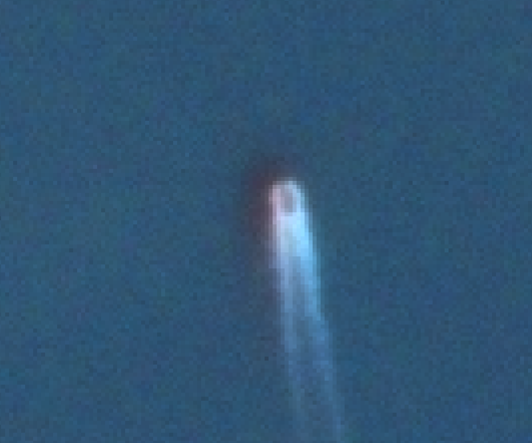}
        \caption{Simulated raw product}
        \label{fig:raw_granule}
      \end{subfigure}
      \hfill
      \caption{Dataset comparison of the same scene. \textit{Maxar Products © 2011-2024 Maxar Technologies.}}
      \label{fig:data_compare}

\end{figure}      
}

\newcommand{\FigClassStats}{%
\begin{figure*}[htbp]
    \centerline{\includegraphics[width=1.8\columnwidth]{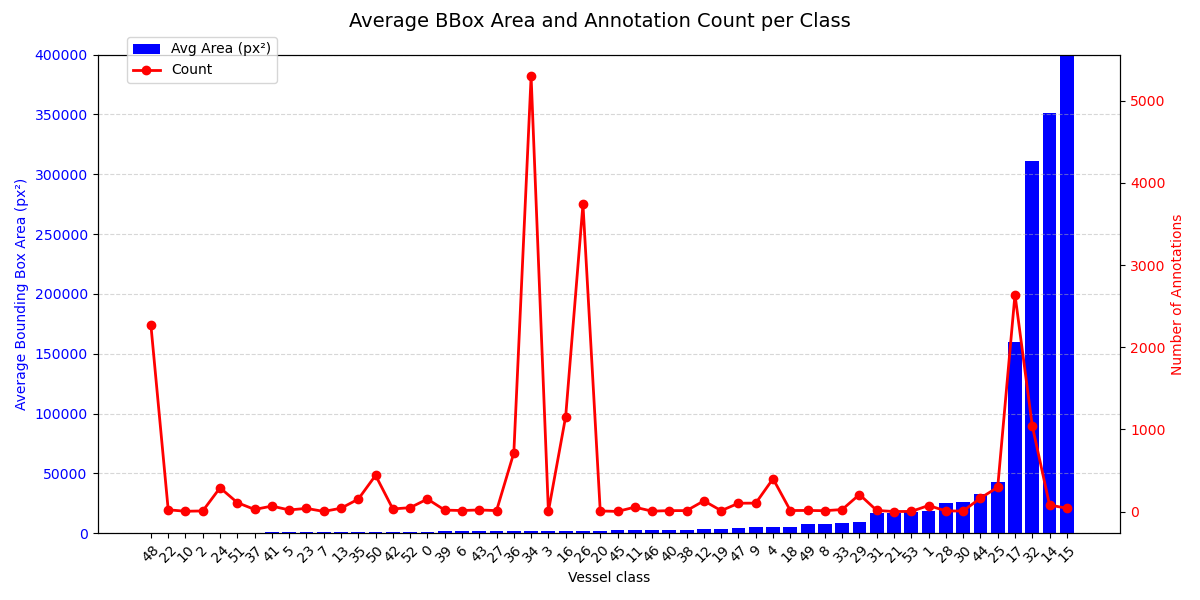}}
    \caption{Vessels area and occurence per class}
    \label{fig:class_stats}
\end{figure*} 
}

\newcommand{\FigVessels}{%
\begin{figure}[!tbp]
 \begin{subfigure}[b]{\columnwidth}
 \centering
    \includegraphics[width=\textwidth]{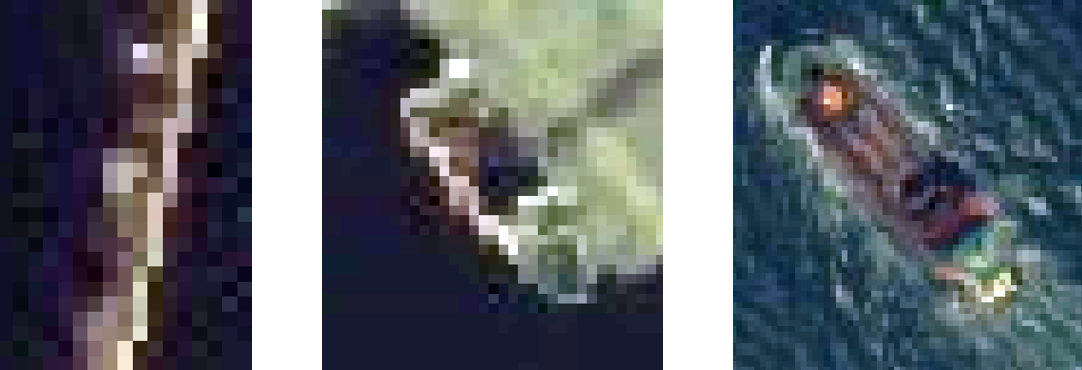}
    \caption{Sailing ship (cls 48), fishing vessel (cls 34) and bulk carrier (cls 50)}
    \label{fig:small_vessels}
  \end{subfigure}
\hfill
 \begin{subfigure}[b]{\columnwidth}
 \centering
    \includegraphics[width=\textwidth]{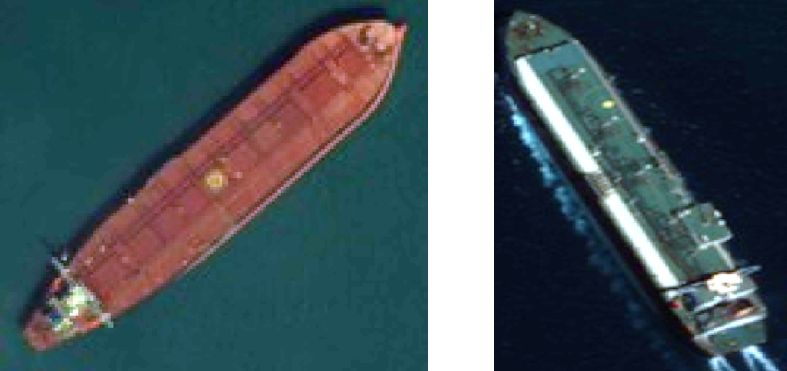}
    \caption{Ore carrier (cls 15) and LNG carrier (cls 14)}
    \label{fig:large_vessels}
  \end{subfigure}
  \hfill
\caption{Small and large vessels comparison. \textit{Maxar Products © 2011-2024 Maxar Technologies.}}
\label{fig:vessels}

\end{figure}     
}

\newcommand{\FigMap}{%
\begin{figure}[!tbp]
  \begin{subfigure}[b]{0.95\columnwidth}
    \includegraphics[width=\textwidth]{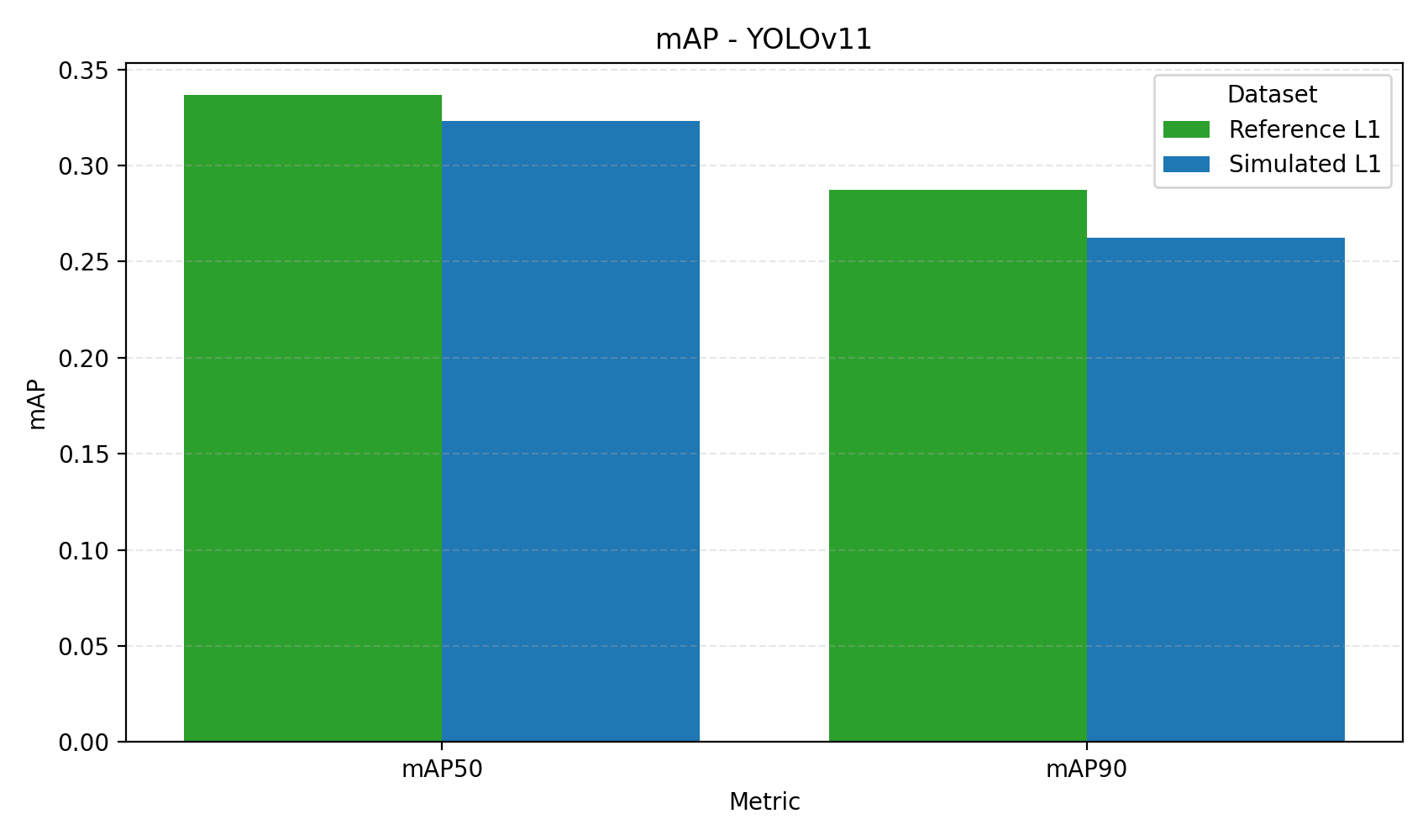}
    \caption{Detection results with a YOLOv11n model}
    \label{fig:map_yolo11}
  \end{subfigure}
  \hfill
  \begin{subfigure}[b]{0.95\columnwidth}
    \includegraphics[width=\textwidth]{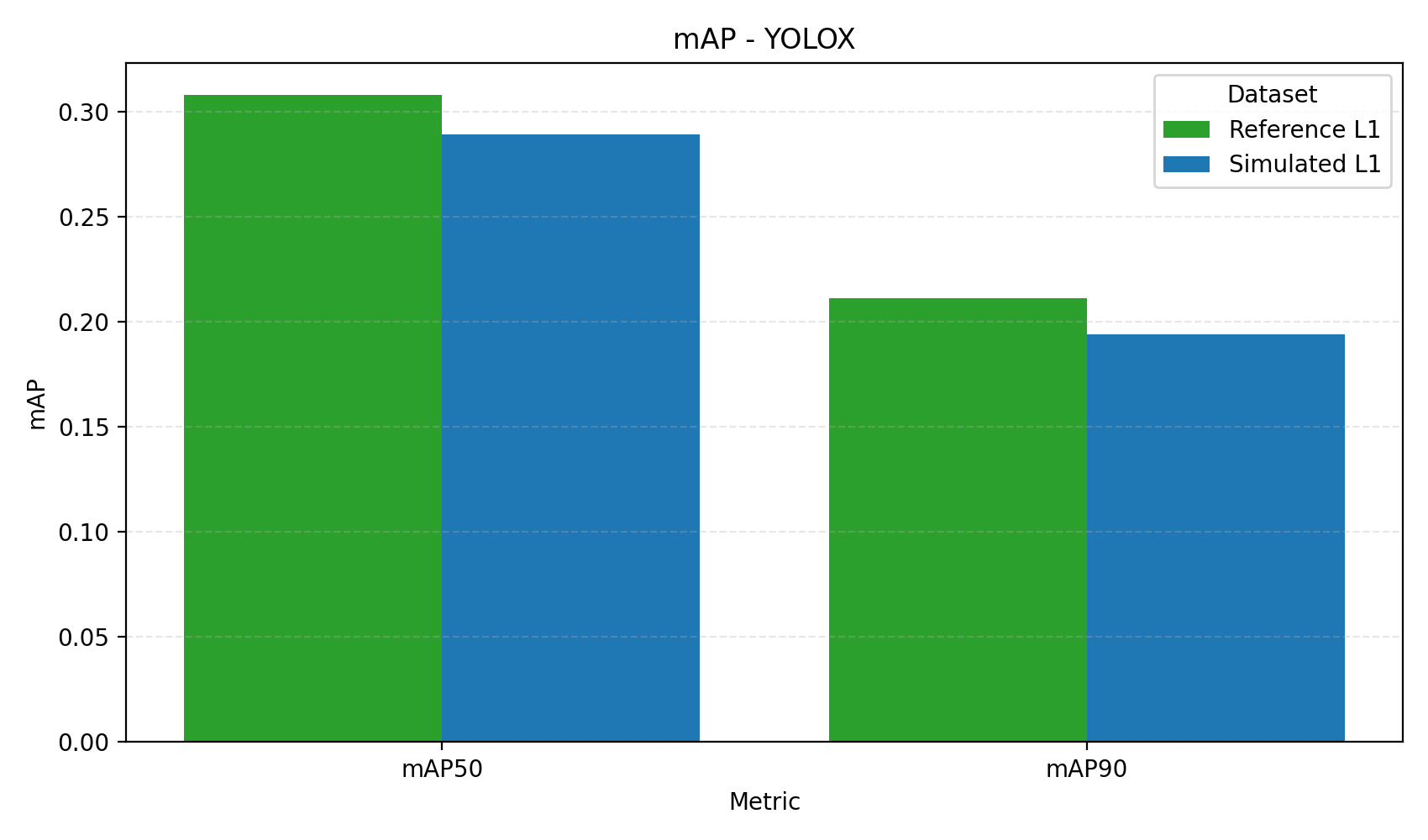}
    \caption{Detection results with a YOLOX-S model}
    \label{fig:map_yolox}
  \end{subfigure}
  \caption{mAP scores on reference L1 and simulated $\tilde{L1}$ datasets}
  \label{fig:map}
\end{figure}  
}

\newcommand{\FigDetectionExample}{%
\begin{figure}[htbp]
    \centerline{\includegraphics[scale=0.75]{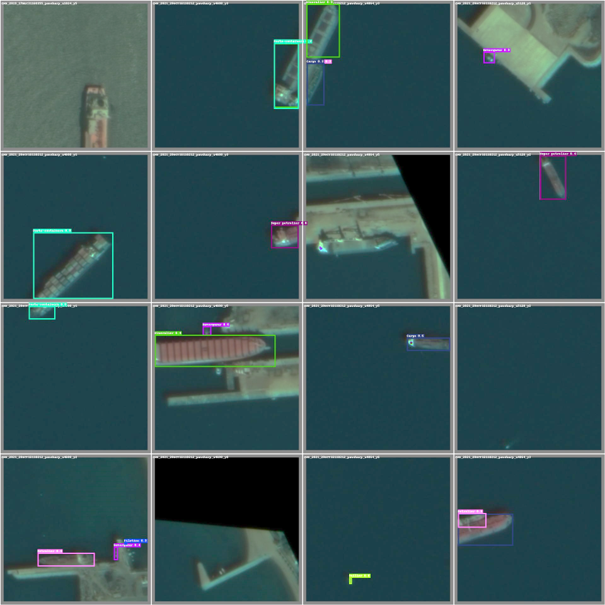}}
    \caption{Mosaic detection on simulated $\tilde{L1}$ dataset with YOLOv11n. \textit{Maxar Products © 2011-2024 Maxar Technologies.}}
    \label{fig:detection_example}
\end{figure} 
}

\newcommand{\FigF}{%
\begin{figure}[!tbp]
      \begin{subfigure}[b]{0.95\columnwidth}
        \includegraphics[width=\textwidth]{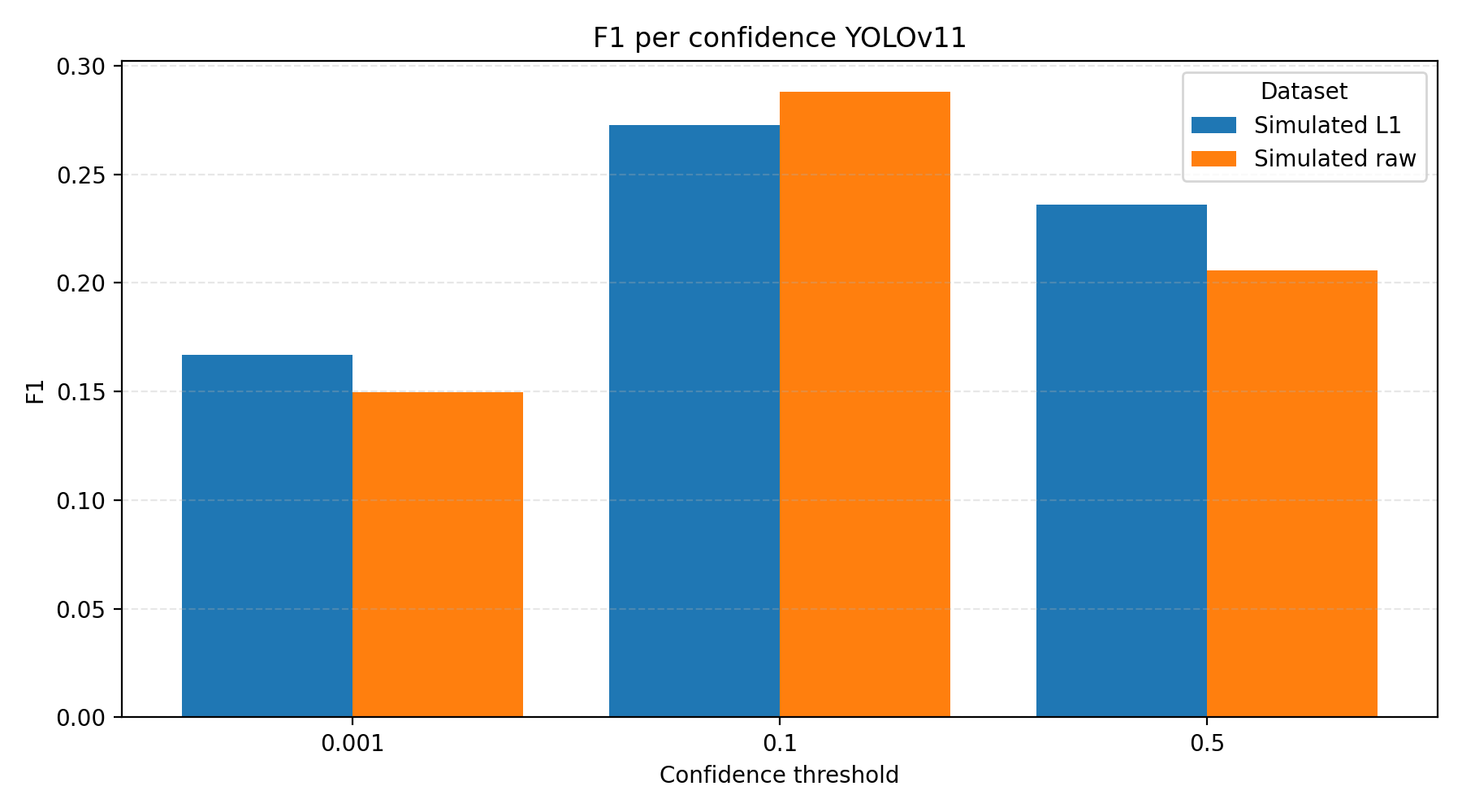}
        \caption{F1 scores with a YOLOv11n model}
        \label{fig:f1_yolo11}
      \end{subfigure}
      \hfill
      \begin{subfigure}[b]{0.95\columnwidth}
        \includegraphics[width=\textwidth]{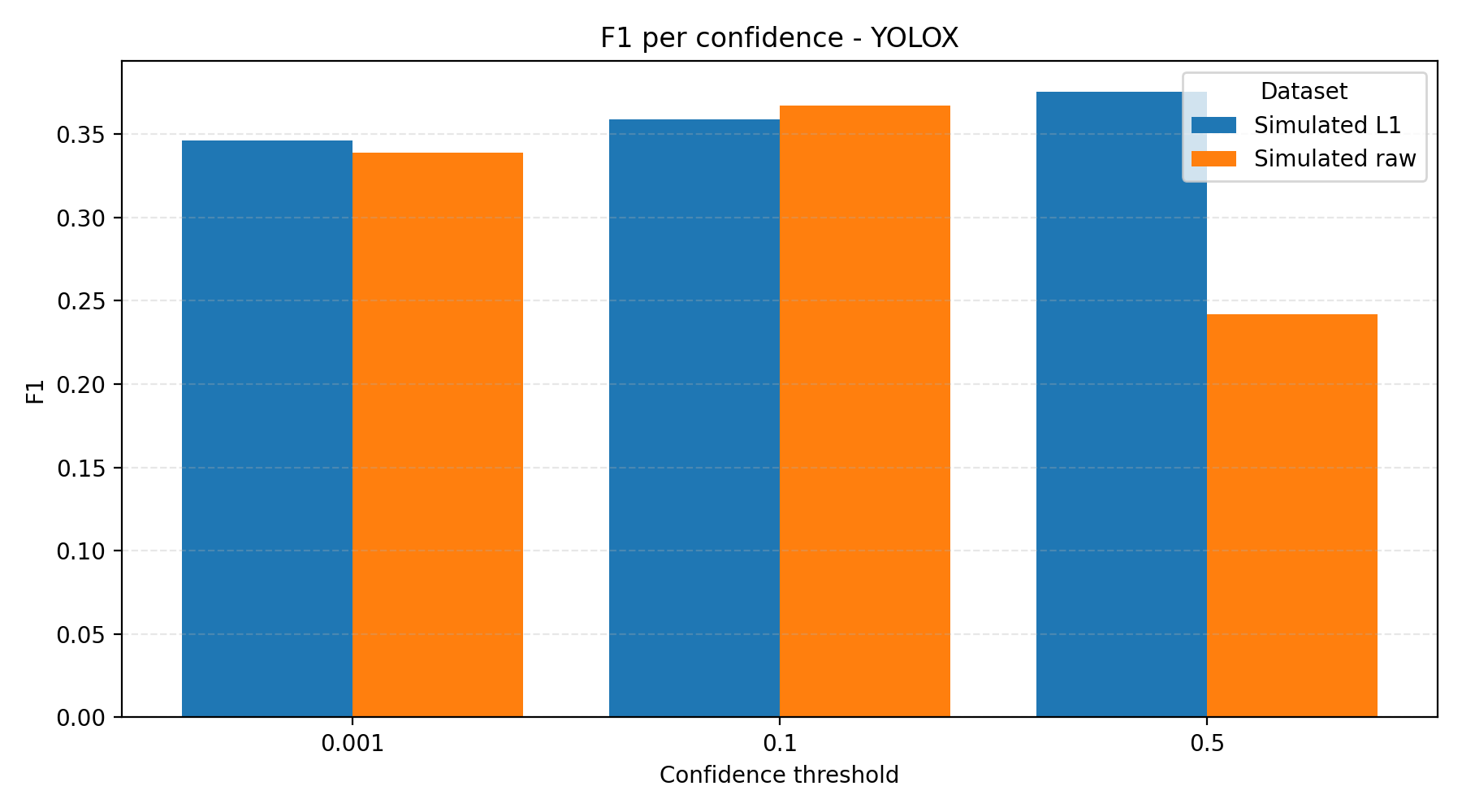}
        \caption{F1 scores with a YOLOX-S model}
        \label{fig:f1_yolox}
      \end{subfigure}
      \caption{F1 scores on $\tilde{L1}$ and raw datasets}
      \label{fig:f1}
    \end{figure}  
}

\newcommand{\FigXplique}{%
\begin{figure}[!tbp]
     \begin{subfigure}[b]{0.47\columnwidth}
     \centering
        \includegraphics[width=\textwidth]{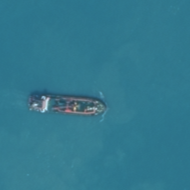}
        \caption{Reference L1 product}
        \label{fig:L1_product}
      \end{subfigure}
    \hfill
     \begin{subfigure}[b]{0.47\columnwidth}
     \centering
        \includegraphics[width=\textwidth]{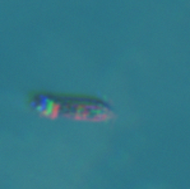}
        \caption{Simulated raw product}
        \label{fig:raw_product}
      \end{subfigure}
      \hfill
      \begin{subfigure}[b]{0.47\columnwidth}
        \includegraphics[width=\textwidth]{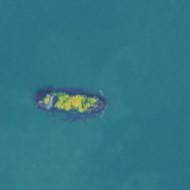}
        \caption{Feature attribution on reference L1 product}
        \label{fig:xplique_l1}
      \end{subfigure}
      \hfill
      \begin{subfigure}[b]{0.47\columnwidth}
        \includegraphics[width=\textwidth]{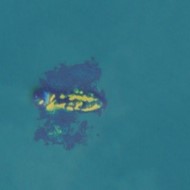}
        \caption{Feature attribution on simulated raw product}
        \label{fig:xplique_raw}
      \end{subfigure}
      \caption{Feature attribution maps using Xplique \cite{xplique}. \textit{Maxar Products © 2011-2024 Maxar Technologies.}}
      \label{fig:xplique}
    \end{figure}
}
\newcommand{\TabDatasetDescription}{%
\begin{table}[tb]
  \caption{Dataset description}\label{tab:dataset_description}
  \centering
  \begin{tabularx}{0.95\columnwidth}{|Y|Y|S[table-format=4.0]|S[table-format=3.0]|S[table-format=2.0]|}
    \rowcolor{black!5}\hline
    \textbf{Patch size} &
    \textbf{Resolution} &
    \textbf{\# train} &
    \textbf{\# val} &
    \textbf{\# classes} \\
    \hline
    \SI{256 } {x} \SI{256}{\px} & \SI{2.0}{m} & 2716 & 306 & 53 \\
    \hline
  \end{tabularx}
\end{table}
}

\newcommand{\TabSimulationParameters}{%
\begin{table}[tb]
  \caption{Simulation parameters}\label{tab:simulation_parameters}
  \centering
  \setlength{\tabcolsep}{6pt}
  \renewcommand{\arraystretch}{1.15}

  \begin{tabular}{|c|c|c|c|c|}
    \hline
    \rowcolor{black!5}
    \textbf{Set} &
    \makecell[c]{\textbf{GSD } \\ \textbf{src / target}} &
    \makecell[c]{\textbf{MTF} \\ \textbf{@Nyquist}} &
    \makecell[c]{\textbf{(Lum., SNR) ref.}\\($W/m^{2}/sr/\si{\micro\metre} $, dB)} &
    \makecell[c]{\textbf{Band} \\ \textbf{offsets}} \\
    \hline
    L1  & 0.5 m / 2.0 m & 0.25 &
    \makecell[c]{(25, 80);\\(100, 170)} & 
    None \\
    \hline
    Raw & 0.5 m / 2.0 m & 0.05  &
    \makecell[c]{(25, 50);\\(100, 110)} &
    [1,4] px \\
    \hline
  \end{tabular}
\end{table}
}
\newcommand{\TabGlobalRestults}{%
 \begin{table}[htbp]
            \centering
            \caption{Detection results across datasets, models, and confidence thresholds}
            \label{tab:global_results}
            \setlength{\tabcolsep}{6pt}
            \renewcommand{\arraystretch}{1.15}
            \begin{tabular}{|c|c|c|c|c|c|c|}
            \hline
            \rowcolor{black!5}
            \textbf{Set} & \textbf{Model} & \makecell[c]{\textbf{Conf.} \\ \textbf{levels}} & \textbf{mAP50} & \textbf{mAP90} & \textbf{F1} & \makecell[c]{\textbf{IoU} \\ \textbf{on TP}} \\
            \hline
            \multirow{6}{*}{\makecell[c]{ref.\\L1}}
             & \multirow{3}{*}{YOLOv11n} & 0.001 & \multirow{3}{*}{0.3366} & \multirow{3}{*}{0.2873} & 0.1019 & 0.8707 \\ \cline{3-3}\cline{6-7}
             &                           & 0.1   &                        &                        & 0.3323 & 0.8916 \\ \cline{3-3}\cline{6-7}
             &                           & 0.5   &                        &                        & 0.3559 & 0.9239 \\ \cline{2-7}
             & \multirow{3}{*}{YOLOX-S}   & 0.001 & \multirow{3}{*}{0.3081} & \multirow{3}{*}{0.2115} & 0.3229 & 0.7882 \\ \cline{3-3}\cline{6-7}
             &                           & 0.1   &                        &                        & 0.3996 & 0.8246 \\ \cline{3-3}\cline{6-7}
             &                           & 0.5   &                        &                        & 0.3805 & 0.8279 \\ \hline
            \multirow{6}{*}{$\tilde{L1}$}
             & \multirow{3}{*}{YOLOv11n} & 0.001 & \multirow{3}{*}{0.3230} & \multirow{3}{*}{0.2624} & 0.1667 & 0.8324 \\ \cline{3-3}\cline{6-7}
             &                           & 0.1   &                        &                        & 0.2727 & 0.8566 \\ \cline{3-3}\cline{6-7}
             &                           & 0.5   &                        &                        & 0.2359 & 0.8862 \\ \cline{2-7}
             & \multirow{3}{*}{YOLOX-S}   & 0.001 & \multirow{3}{*}{0.2895} & \multirow{3}{*}{0.1942} & 0.3463 & 0.7876 \\ \cline{3-3}\cline{6-7}
             &                           & 0.1   &                        &                        & 0.3590 & 0.8096 \\ \cline{3-3}\cline{6-7}
             &                           & 0.5   &                        &                        & 0.3754 & 0.8182 \\ \hline
            \multirow{6}{*}{Raw}
             & \multirow{3}{*}{YOLOv11n} & 0.001 & \multirow{3}{*}{0.3182} & \multirow{3}{*}{0.2584} & 0.1495 & 0.8352 \\ \cline{3-3}\cline{6-7}
             &                           & 0.1   &                        &                        & 0.2879 & 0.8675 \\ \cline{3-3}\cline{6-7}
             &                           & 0.5   &                        &                        & 0.2059 & 0.9128 \\ \cline{2-7}
             & \multirow{3}{*}{YOLOX-S}   & 0.001 & \multirow{3}{*}{0.2667} & \multirow{3}{*}{0.1876} & 0.3387 & 0.7898 \\ \cline{3-3}\cline{6-7}
             &                           & 0.1   &                        &                        & 0.3671 & 0.8101 \\ \cline{3-3}\cline{6-7}
             &                           & 0.5   &                        &                        & 0.2418 & 0.8860 \\ \hline
            \end{tabular}
        \end{table}
}

\iftrue

    \newcommand{\todo}[1]{(to do $\rightarrow$ \textbf{#1})}

\else

    \newcommand{\todo}[1]{}

\fi

\begin{document}

\title{Explaining raw data complexity to improve satellite onboard processing}

\author{
    \IEEEauthorblockN{
        Adrien Dorise \IEEEauthorrefmark{1}\IEEEauthorrefmark{2},
        Marjorie Bellizzi \IEEEauthorrefmark{2},
        Adrien Girard \IEEEauthorrefmark{2},
        Benjamin Francesconi \IEEEauthorrefmark{2},
        Stéphane May \IEEEauthorrefmark{1}
    }

    \IEEEauthorblockA{\IEEEauthorrefmark{1}CNES, Toulouse, FRANCE\\
    Emails: \{adrien.dorise, stephane.may\}@cnes.fr}

    \IEEEauthorblockA{\IEEEauthorrefmark{2}IRT Saint-Exupéry, Toulouse, FRANCE\\
    Emails: \{marjorie.bellizzi, adrien.girard, benjamin.francesconi\}@irt-saintexupery.com}
}

\maketitle

\begin{abstract}
    With increasing processing power, deploying AI models for remote sensing directly onboard satellites is becoming feasible. However, new constraints arise, mainly when using raw, unprocessed sensor data instead of preprocessed ground-based products. While current solutions primarily rely on preprocessed sensor images, few approaches directly leverage raw data. This study investigates the effects of utilising raw data on deep learning models for object detection and classification tasks. We introduce a simulation workflow to generate raw-like products from high-resolution L1 imagery, enabling systemic evaluation. Two object detection models (YOLOv11n and YOLOX-S) are trained on both raw and L1 datasets, and their performance is compared using standard detection metrics and explainability tools. Results indicate that while both models perform similarly at low to medium confidence thresholds, the model trained on raw data struggles with object boundary identification at high confidence levels. It suggests that adapting AI architectures with improved contouring methods can enhance object detection on raw images, improving onboard AI for remote sensing.

\end{abstract}


\section{Introduction}
    The computational resource requirements of AI algorithms have traditionally limited the use of artificial intelligence (AI) in space applications. As a result, most AI processing has been performed on the ground.
    However, with new components being tested for potential use in space \cite{versal_heavy_ion}, real-time data processing and analysis is becoming feasible, enabling artificial intelligence models directly onboard the satellite. Recent projects have explored the potential of onboard AI applications \cite{irma_imagini, scannernet, onboard_oil_spill_detection} for remote sensing, including the OPS-SAT project, which demonstrated the feasibility of running modern AI algorithms on satellite hardware \cite{opssat_tensorflow, opssat_meoni}. Processing data directly onboard offers significant advantages in terms of mission efficiency, leading to increased reactivity, better resource management, and bandwidth saving \cite{remote_sensing_review}.
    \\
    \\
    Nevertheless, most existing applications rely on preprocessed images similar to ground Level-1 (L1) products. These images undergo advanced post-processing techniques, such as coregistration, geometric and radiometric corrections, which can be computationally expensive for embedded satellite systems.
    In contrast, raw sensor outputs are noisy, deregistered, and affected by artefacts \cite{pyraws_maritime1}. We refer to these as raw products. The European Space Agency's $\phi-lab$ has made significant contributions in using directly raw images as input for AI models, by proposing an onboard coarse coregistration package called PyRawS \cite{pyraws_thraws}, and creating publicly available datasets based on Sentinel-2 for evaluation \cite{pyraws_maritime2}. 
    \\
    \\
    Despite these advances, little research has examined how raw high-resolution satellite images affect AI performance, primarily due to the lack of suitable datasets. In this study, we address this gap by developing a simulation workflow that generates raw-like products from L1 imagery. Our approach combines panchromatic (PAN) and multispectral (XS) products, applies sensor-like degradations, and reconstructs both L1 and raw datasets. Using vessel detection as a case study, we train and compare AI models to analyse how the degradation impacts performance. From there, we will leverage explainable AI techniques \cite{explainableAI(XAI), xplique}, which reveal how degradations influence feature attribution and boundary localisation. By analysing how models perceive perturbations induced by raw data, we aim to gain insights into the significant challenges AI models encounter when processing raw images.
    \\
    \\
    This study is divided into the following sections. Section 2 provides an overview of the material and methods used, including a thorough description of the data simulator developed during this study, as well as a description of the datasets and the trained AI architectures.
    Section 3 describes the experimental setup developed to assess the impact of raw data on a vessel detection use case. Section 4 presents our experiments, comparing the performance of models trained on the reference L1 images, simulated $\tilde{L1}$ images and simulated raw images. Section 5 discusses the validity of our simulation tool and the effects of raw data. Finally, section 6 concludes the paper by summarising our key contributions and outlining future research directions.

\section{Materials and methods}

    In a related work, Del Prete et al. [3] focused on vessel detection and classification using Sentinel-2 (10–80m GSD) and Venµs (5m GSD) imagery. In our study, we address a similar task but leverage significantly higher-resolution data. High-resolution bundle products typically consist of a single panchromatic (PAN) image and multiple lower-resolution multispectral (XS) images. Since the focus of this study is the impact of raw data on detection performance, it is necessary to simulate a raw equivalent image from these products. For this purpose, a simulation workflow is developed.
    
    \subsection{Simulation workflow}   
    To obtain high-resolution colour images, a common approach when working with PAN and XS data is to restore the PAN image and combine it with the XS images to generate a high-resolution pansharpened multispectral image \cite{pleiades_restoration}. In this work, we developed a custom processing workflow to simulate both L1 and raw products from the original PAN and XS images. The proposed workflow is structured as follows:
    \begin{itemize}
        \item First, both original PAN and XS images are degraded to match a raw product similar to the sensor's output, creating the $PAN_{raw}$ and $XS_{raw}$ products.
        \item The $PAN_{raw}$ product is restored using an embedded neural network \cite{EDSR} with an upscale factor of 1, creating a $PAN_{restored}$ product. 
        \item A pansharpening algorithm is performed on $XS_{raw}$ and $PAN_{restored}$ to create a $PANSHARP_{restored}$ product.
        \item Finally, the pansharpened product is divided into patches that can be input into a deep learning model for detection.
    \end{itemize}
    
    To create a restored pansharpened product, the whole workflow is executed; however, the PAN restoration step is skipped when building a raw pansharpened product. The complete workflow can be visualised in Fig. \ref{fig:image_workflow}.
    
    \FigImageWorkflow

    \subsubsection{Degradation}
    
        \FigSensorSimuFlowchart

        The degradation is addressed by a sensor simulator tool. 
        This simulator reverses the restoration process, transforming an L1 product into a degraded raw product (L0). A description of the available degradations and their processing order is provided in Fig. \ref{fig:sensor_simu_flowchart}. In addition, Fig. \ref{fig:simu} illustrates an example of the results for some simulator functions.
        The noise is added by following a signal-dependent noise model of the form:
        \begin{equation}
            \sigma^2 = \alpha L+\beta
        \end{equation}
        where $L$ is the luminance values of the image and $\alpha$, $\beta$ are parameters describing how the noise scales with luminance.
        The noise parameters are computed by using the luminance and SNR of two reference points characterising the targeted sensor: $(L_{dark},SNR_{dark}), (L_{bright},SNR_{bright})$.
        \\
        To maintain full control over the degradation process, particularly during MTF-related effects, the original product is downsampled by a factor of x4.

        \FigSimu

    \subsubsection{Restoration}
        Traditional satellite image processing chains rely on complex physical models to enhance the quality of raw products. In this work, we employ neural networks to restore raw images. The super-resolution model Enhanced Deep Super-Resolution network (EDSR) \cite{EDSR} is used for this task. During the training phase, it was fed with raw images created by the degradation process as input, and downscaled L1 images as output. This step is skipped when simulating a raw image.

    \subsubsection{Pansharpening}
       The pansharpening process involves merging a high-resolution panchromatic image and an upsampled low-resolution multispectral image to create a high-resolution multispectral image. In this work, the Brovey method is applied to the XS product to generate the pansharpened products. For a resampled $N$-band multispectral product $\{M_i(\mathbf{x})\}_{i=1}^N$ and a PAN image $P(\mathbf{x})$, the fused band $F_i$ is
        \begin{equation}
        F_i(\mathbf{x}) \;=\; 
        \frac{M_i(\mathbf{x})}{\displaystyle \sum_{j=1}^{N} w_j\,M_j(\mathbf{x})}\; P(\mathbf{x}),
        \label{eq:brovey}
        \end{equation}
        where $w_j$ are optional weights. 

    \subsubsection{Tiling}
        The tiling algorithm was first implemented using a straightforward grid division of the entire granule. The division is arbitrary and depends on the desired output size. However, we discovered that many of the vessels of interest were split into different patches, making detection more difficult.
        \\
        To address this issue, we shifted the focus to the vessels themselves. Each vessel must appear entirely in at least one patch. To ensure this, a new patch is generated for each vessel in the product. The vessel is centered within the patch, with a random offset in X and Y directions of 30\% of the patch size to avoid overfitting during training.

    \subsection{Datasets}
        For this work, we rely on a novel and custom-built database of HR bundle products specifically designed to develop and evaluate AI-based ship detection algorithms. This unique and high-quality dataset was created within the IRT framework and comprises 47 high-resolution scenes (30–50cm GSD) totalling over 24,000 annotated ships across 53 vessel classes, including small boats, military vessels, and commercial cargo ships. The imagery originates from MAXAR Standard (2A) products, which include 1 PAN and R, G, B bands (at a resolution 4 times lower than PAN) with standard radiometric and geometric corrections. All annotations were performed by expert photointerpreters from GEO4I.
        \\
        In this study, the HR bundle images serve as reference (source) data for simulating two types of downstream products using the simulation workflow:
        \begin{itemize}
            \item Simulated raw-like images representative of a sensor’s native output, to evaluate detection algorithms tailored to raw data. These are used to create the simulated raw datasets.
            \item Simulated Level-1 images (corrected, restored, and pansharpened), to assess detection performance under traditional processing workflows. These are used to create the simulated $\tilde{L1}$ dataset.
        \end{itemize}

         The parameters used for the simulation are available in Table \ref{tab:simulation_parameters}.
        
        \TabSimulationParameters

        In addition, the original RGB images are tiled to create the reference L1 dataset. The description of the datasets is available in Table. \ref{tab:dataset_description}.
        In the comparison of the three datasets shown in Fig. \ref{fig:data_compare}, we observe a slight modification in the colours of the simulated $\tilde{L1}$ dataset. Experiments will disclose if these images can be used for effective comparison. Furthermore, it is visually apparent that degradation is occurring in the simulated raw product.

        \TabDatasetDescription       
        
        \FigDataCompare

    \subsection{Models architecture}

        Object detection and classification can be done in two distinctive ways: performing two-stage detection, where the model first identifies regions of interest and then performs detection and classification within these regions \cite{RCNN, faster_RCNN}, or single-stage detection, where the whole image is used for object detection and classification \cite{vessel_detection_yolo11_sentinel2}. Although two-stage detectors usually yield better results, they are often slower than a single-stage detector. Since our objective is real-time onboard processing, we focus exclusively  on small single-stage detection models. Therefore, we selected YOLOv11n (2.6 million parameters) \cite{yolo11} and YOLOX-S (9.0 million parameters) \cite{yolox} architectures. Previous experiments have demonstrated the successful deployment of these YOLO models on an AMD Versal board \cite{CIAR}, confirming their compatibility with embedded hardware.

    \section{Experiments}
        Two hypotheses are to be tested in this study. First, we want to assess the validity of our simulation workflow. It is done by comparing the models' performance on the reference L1 and simulated $\tilde{L1}$ dataset. Indeed, a loss in performance would indicate that the simulated data does not faithfully represent the original information.
        Then we want to evaluate the impact raw images have on detection models. The goal is to verify the possibility of using raw data for vessel detection use cases. By doing so, it would enable the onboard processing of the satellite's sensor output.

        \subsection{Small objects removal}
            Similar to the work of Goudemant et al. \cite{CIAR}, the first experiments included all 53 classes available in the dataset. However, the results showed that the models were unable to classify small vessels accurately. Indeed, the original work performed vessel detection on 0.5m resolution images. However, due to the image processing performed, we are limited to a resolution of 2.0 m.
            \\
            Fig. \ref{fig:class_stats} illustrates the distribution of each class by area and the number of occurrences. We observe substantial disparities in vessel size. Consequently, we decided to retain only the six largest vessel classes from our dataset.
            \\
            Sampled examples of large and small vessels, shown in Fig. \ref{fig:vessels}, helped clarify that the information given at 2.0 m resolution is insufficient to classify all vessel types accurately.
            
            \FigClassStats
            \FigVessels
    
        \subsection{Model training}
            Each of the two architectures (YOLOv11n and YOLOX-S) was trained at least three times on the three available datasets (original L1, simulated $\tilde{L1}$ and simulated raw). 
            To ensure a fair comparison, the same model architecture and identical hyperparameters were used across the three datasets. By doing so, it is possible to compare the performance of each model on the datasets accurately. Indeed, the difference in performance would stem solely from the datasets, without any external factors that could impede the results. Of course, the randomness of the training could alter the outcome of a run from another. That is why the models were trained multiple times with similar settings to avoid outliers.
            The models were trained from scratch for 300 epochs using a warm cosine scheduler, along with data augmentation techniques such as flipping, rotating, and HSV adjustments. The images were upsampled from 256x256 pixels to 640x640 pixels.

        \subsection{Metrics  \label{sub:metrics}}
            Two types of metrics were used in this study.
            The first type of metric is related to the detection task: 
            \begin{itemize}
                \item Mean Average Precision (mAP): evaluate the global performance of the model.
                \item F1 score across different confidence thresholds: measure how a model performs when detecting complex objects.
                \item Intersection Over Union (IoU) on correctly predicted samples: reflects the localisation accuracy of the predictions.
            \end{itemize}       
            
            The second metric type is based on Xplique, an explainability tool that generates maps highlighting the key input regions a vision model uses to make predictions \cite{xplique}. By comparing these maps for both $\tilde{L1}$ and raw datasets, we can visualise how the degradations impact the model's performance.

\section{Results}
    The results of all the experiments are summarised in Table \ref{tab:global_results}. In this section, we present the results of the simulation workflow validation and the comparison between the simulated $\tilde{L1}$ and the simulated raw datasets.

    \TabGlobalRestults

    \subsection{Validation of the simulation workflow}

        Fig. \ref{fig:map} compares the mAP50 and mAP95 scores for YOLOv11n and YOLOX-S models trained on the reference L1 dataset and on the simulated $\tilde{L1}$ dataset. Regarding YOLOv11n, performance decreases slightly from an mAP50 of 33.6\% to 32.3\% when trained on simulated $\tilde{L1}$ data. A similar drop can be observed when examining the YOLOX-S model, with the mAP score decreasing from 30.81\% to 28.9\%. These results indicate that although models trained on simulated data underperform compared to those trained on reference L1 data, the simulated $\tilde{L1}$ data remains close to the reference L1 products to be used in subsequent experiments.
        \\\\
        Qualitative examples of vessel detections on $\tilde{L1}$ datasets are shown in Fig. \ref{fig:detection_example}. These confirm that the models can detect vessels across simulated L1 images.  
    
        \FigMap
        
        \FigDetectionExample

    \subsection{Impact of raw products}

    When examining the mAP50 and mAP95 scores for the simulated L1 and simulated raw datasets from Table \ref{tab:global_results}, we observe a slight deterioration in the performance of models trained on raw data.
    \\\\
    To analyse the effect of raw data in more detail, F1 scores were evaluated across multiple confidence thresholds. Fig. \ref{fig:f1} displays the evolution of the F1 score for both YOLOv11n and YOLOX-S on reference L1 images and simulated raw images. The thresholds were chosen to represent low/medium/high confidence. It demonstrated that for low to medium confidence levels, the models performed similarly. However, at higher thresholds, the performance of the models trained on raw data deteriorates more rapidly.

    \FigF

    In addition to these results, explainability maps generated with the Xplique library are shown in Fig. \ref{fig:xplique}, highlighting areas of interest in the image used by the model to make predictions. We observe that the model struggles to identify the vessel boundaries when trained on raw data, whereas it can extract the vessel's shape with greater precision using the original product.

    \FigXplique

    \section{Discussions}
        Two main insights can be drawn from the results. First, the simulation workflow generates data sufficiently close to the original L1 products for evaluating AI models. Although a slight performance drop is observed, the dataset preserves most of the relevant information and can be used for reliable experiments.
        \\\
        Second, the comparison between simulated $\tilde{L1}$ and raw data indicates that degradation primarily affects the robustness of predictions at high confidence thresholds. Indeed, the rapid decline in F1 scores indicates that, while models trained on raw data can detect and classify vessels, they struggle to provide consistent, high-confidence predictions. This suggests that processing raw data is especially challenging when handling edge cases and identifying complex objects.
        \\\\
        The explainability analysis helps clarify this behaviour. Models trained on raw data struggle to clearly identify the boundaries of the object, resulting in less reliable predictions. Since accurate boundary localisation is essential for robust predictions, this explains part of the drop in robustness. It represents a novel approach to building reliable models that operate effectively with sensor output. Nevertheless, further investigations using larger and more diverse samples are required to validate these findings.

\section{Conclusion}

    In this study, we analysed the impact of raw (L0) images on deep learning models trained for object detection. Using a dataset of annotated L1 vessel products, we implemented a simulation workflow that generates equivalent raw and $\tilde{L1}$ images from processed high-resolution satellite images. The processing chains include a degradation tool that attempts to emulate onboard sensors, as well as a neural network-based restoration mechanism. A vessel detection use case is used, and two models were then trained separately on the reference L1 images and on the simulated L1 and raw datasets. Their performance was evaluated using detection metrics and explainability visualisations. The results of the simulated workflow are satisfactory, and the images can be effectively used to evaluate model performance. In addition, the experiments on raw data suggest that while object detection capabilities remain largely intact, raw image degradation significantly affects the precision and confidence of predictions.
    \\
    \\
    These results provide initial insights into how models behave when processing raw data. These indications will be crucial in identifying the most impactful degradations. Future work includes further investigation of models' predictions using explainable tools. Moreover, to fully leverage high-resolution imagery, work will be conducted to classify smaller vessels efficiently. Finally, work is ongoing to develop novel AI architectures designed to mitigate these effects and evaluate their performance on embedded hardware. These advancements will bring us closer to efficiently deploying AI models directly on board satellites, thereby enhancing autonomous remote sensing capabilities.

\section*{Acknowledgment}
    
    The work presented in this article is conducted as part of the IRMA project (“Image processing for a Responsive Mission with AI”) at IRT Saint-Exupéry. This project seeks to develop technological components based on artificial intelligence for mission planning and data processing, both onboard and on the ground, with the aim of addressing the evolving needs in the Earth observation market, such as the transformation of raw data into interpreted information and the reactive management of satellite constellations. IRMA also seeks to demonstrate the effectiveness of these technologies through system loop demonstrations, whether in the laboratory or in orbit. 
    The authors would like to thank the industrial and academic partners of the project: Thales Alenia Space, Activeeon, Geo4i, JoliBrain, University Côte d’Azur and CNES.

\bibliographystyle{abbrv}
\bibliography{biblio.bib}	

\end{document}